\documentclass{article}
\PassOptionsToPackage{numbers,sort&compress}{natbib}
\usepackage[preprint]{neurips_2026}
\usepackage{tcolorbox} 
\tcbuselibrary{skins}
\usepackage[utf8]{inputenc}
\usepackage[T1]{fontenc}
\usepackage{hyperref}
\usepackage{url}
\usepackage{booktabs}
\usepackage{amsfonts}
\usepackage{amsmath}
\usepackage{amssymb}
\usepackage{nicefrac}
\usepackage{microtype}
\usepackage{graphicx}
\usepackage{doi}
\usepackage{xcolor}
\usepackage{colortbl}
\usepackage{multirow}
\usepackage{subcaption}
\usepackage{algorithm}
\usepackage{algorithmic}
\usepackage{enumitem}
\tcbuselibrary{skins,breakable,listings}
\usepackage{tikz}
\usepackage{fontawesome5}

\tcbuselibrary{breakable}

\usepackage[textsize=scriptsize, tickmarkheight=0pt]{todonotes}
\usepackage{lipsum}

\newcommand{\RARG}{%
  R%
  \kern-.25em                          
  \raisebox{0.45ex}{A}
  \kern-.18em                          
  R%
  \kern-.22em                          
  \raisebox{-0.15ex}{G}
}

\newcommand{\method}{RARG}
\newcommand{\methodp}{RARG+}
\newcommand{\methodpp}{RARG++}
\newcommand{\grep}{\texttt{grep}}
\newcommand{\rg}{\texttt{rg}}

\newcommand{\lstterm}[2][]{%
  \setlength{\fboxrule}{0.4pt}%
  \setlength{\fboxsep}{2pt}%
  \fcolorbox{gray!50}{gray!8}{%
    \lstinline[basicstyle=\ttfamily\small\color{#1}]{#2}%
  }%
}

\definecolor{heatlow}{HTML}{FDCFCF} 
\definecolor{heatmid}{HTML}{FFFFCC} 
\definecolor{heathigh}{HTML}{C6EFCE} 
\definecolor{heatbest}{HTML}{7BC67E} 
\definecolor{heatgray}{HTML}{F0F0F0} 

\title{A New Role for Relevance: Guiding Corpus Interaction in Agentic Search}

\author{
 Jiangnan Li$^{1,}$\thanks{equal contributions; $^{\dagger}$ corresponding authors}~~, 
 Yuqing Li$^{2,*}$, 
 Mo Yu$^{1,\dagger}$, 
 Jinchao Zhang$^{1}$, 
 Jie Zhou$^{1,\dagger}$ \\
$^{1}$Tencent\ \ \ \ \ $^{2}$IIE-CAS \\
$\{$\texttt{jiangnanli,moyumyu}$\}$\texttt{@tencent.com}, \texttt{liyuqing@iie.ac.cn}
}

\begin{document}
\maketitle
\vspace{-2.0em}
\begin{center}
\faGithub~
\href{https://github.com/LeqsNaN/RARG}
{\texttt{github.com/LeqsNaN/RARG}}

\end{center}
\vspace{1.2em}

\begin{abstract}

Relevance is a query-dependent estimate of whether a document or excerpt contains useful evidence. Existing retrieval agents use relevance to select top-$k$ content, but document relevance alone cannot localize, compose, or verify the evidence required by complex questions. Direct Corpus Interaction (DCI) enables such fine-grained operations through grep-style exploration, but its relevance-agnostic search can expose useful clues late and delay convergence. Recent advances use relevance to narrow the corpus into a working space for interaction. Once interaction begins, however, relevance still does not directly guide which documents grep searches first or distinguish informative excerpts from a broad set of matches to let LLMs see them first. We introduce the \textbf{Relevance-Aware RipGrep Search Agent (\method)}, which turns relevance into an execution prior for corpus interaction. RARG provides coarse-to-fine relevance guidance: it orders documents for sequential \texttt{ripgrep} traversal to expose globally relevant clues earlier, initializes promising entry points with query-relevant paragraphs, and reranks grep matches to surface informative excerpts that document-level ranking may otherwise obscure. Across challenging browse question answering and reasoning-intensive retrieval, RARG improves the accuracy--efficiency frontier over retrieval-based and direct-interaction agents. These results demonstrate that relevance-aware interaction enables faster and more reliable search convergence.
\end{abstract}

\begin{figure*}[ht!]
    \centering
    \includegraphics[width=0.9\linewidth]{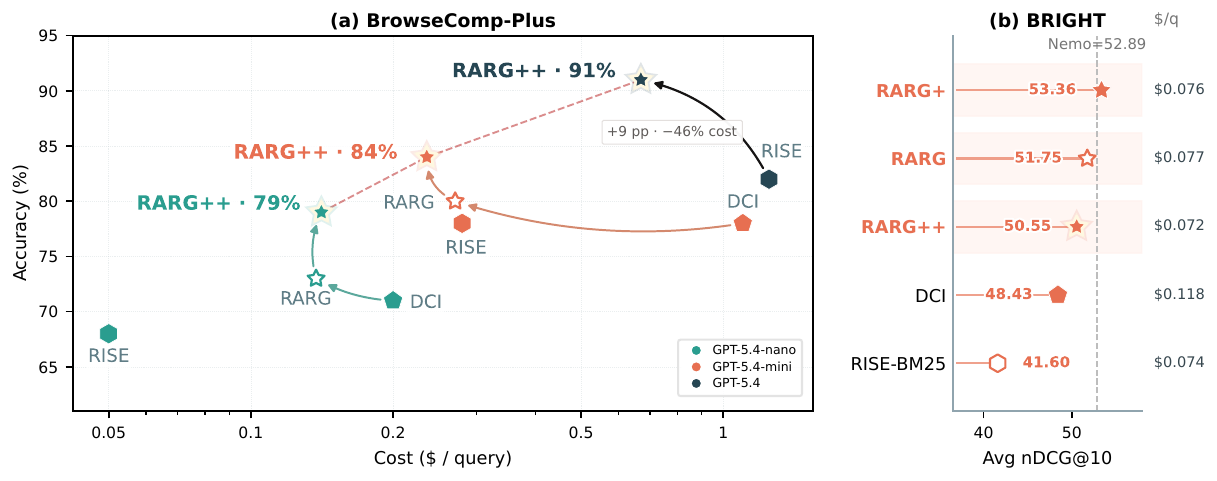}
    \caption{Accuracy/nDCG@10 versus interaction cost (average tool calls) on BrowseComp-Plus and BRIGHT. By turning relevance into an execution prior over \rg~exploration, \method~advances the accuracy--efficiency frontier over retrieval-based and direct-interaction agents.}
    \label{fig:cost}
\end{figure*}

\section{Introduction}

Relevance has long served as the organizing principle of information retrieval. In this paper, we use \emph{relevance} to mean a query-dependent estimate of how likely a document, passage, or matched excerpt is to contain evidence useful for the current information need. Modern dense retrievers, which are widely deployed in RAG~\cite{lewis2020retrieval,gao2023retrieval,singh2025agentic,Mindscape}, instantiate this estimate with embedding similarity~\cite{karpukhin2020dense,Qwen3Emb,wu2026situated}, while sparse retrievers use lexical matching signals~\cite{robertson2009probabilistic}. In either case, relevance is a useful but imperfect prior: it indicates where evidence may be found, not whether the evidence is sufficient, correctly localized, or properly combined with other clues.

Retrieval agents conventionally use this prior to rank the corpus and expose a top-$k$ set of documents or snippets to the language model~\cite{yao2022react,searcho1,searchr1,r1searcher,deepresearcher,webthinker}. This interface is scalable, but it conflates document relevance with evidence utility. A relevant document may hide the decisive fact in a small span omitted by snippet truncation~\cite{liu2024lost}, while in multi-hop search its usefulness may emerge only after an intermediate entity is discovered~\cite{trivedi2023interleaving}. Independent query--document ranking cannot itself localize, connect, compare, or verify such evidence. An agent may therefore retrieve the right documents yet still fail to produce the answer, as observed in prior studies~\cite{DCI}. Relevance can identify promising regions of a corpus, but a ranked top-$k$ view provides limited support for exploiting the evidence within and across them.

Direct Corpus Interaction (DCI) addresses this interface bottleneck by allowing an agent to search raw documents with general-purpose terminal tools such as pattern matching and local file reads~\cite{DCI}. Fine-grained interaction lets the agent combine lexical constraints, follow newly discovered entities, and inspect the exact context surrounding a match. However, DCI largely removes corpus-level relevance guidance: a pattern-matching command scans documents as if all locations were equally promising, so useful clues may appear late or be lost when its output is truncated. The agent can consequently spend many unproductive interactions exploring fruitless search angles before reaching a useful anchor. We refer to how quickly and reliably these iterative interactions narrow an open-ended search into verified evidence as \emph{search convergence}. Without a mechanism for prioritizing promising locations, DCI's convergence deteriorates as the corpus grows, leading to rapidly increasing tool use and lower accuracy~\cite{DCI}.

Recent advances, including Pi-Serini, RISE, and DR-DCI, address this scaling problem by using sparse or dense retrieval to construct or expand a bounded candidate space before or during interaction~\cite{PiSerini,RISE,DRDCI}. Despite differences in their interfaces, they share a common strategy: first narrow the corpus to a set of relevant documents, and then let the agent explore this set through selective reads or grep-style operations. This avoids repeated full-corpus scans while preserving richer interaction than a conventional top-$k$ snippet interface. However, relevance remains primarily a mechanism for constructing the space, rather than a fine-grained execution signal for the interactions within it. It does not directly determine both the traversal order of a pattern-matching operation and which local matches remain visible under output truncation.

We argue that relevance should guide corpus interaction, rather than merely select its inputs, to accelerate and stabilize search convergence. This requires relevance at two complementary resolutions. \emph{Global relevance} should determine where a pattern-matching operation searches first, so that promising documents are encountered before less relevant ones. \emph{Local relevance} should determine which matched excerpts become visible when the raw output exceeds the agent's observation budget. This view preserves the high-resolution, compositional interface of DCI while using retrieval scores as an execution prior instead of a hard evidence bottleneck.

Based on this principle, we introduce the \textbf{Relevance-Aware RipGrep Search Agent (\method)}.\footnote{Throughout the paper, \grep~denotes the generic grep-style pattern-matching operation, whereas \rg~denotes the concrete \texttt{ripgrep} command used by our implementation.} \method~ranks documents with an embedding retriever according to an agent-generated query and then makes \rg~traverse them sequentially in that order, so that matches from more relevant documents are exposed earlier---turning document-level relevance into search order rather than top-$k$ content selection. We further study two extensions. \methodp~seeds the agent with a small set of query-relevant paragraphs as an entry point~\cite{MiASignature} from which it can formulate its first precise search. \methodpp~reranks a wider pool of \rg~matches by combining the global retrieval objective with the local pattern-matching focus, so that locally informative excerpts---including evidence in lower-ranked documents---can compete for the agent's limited observation budget. Together, the three levels determine where interaction begins, which documents it traverses first, and which local observations reach the model, helping \rg~reach useful evidence earlier and converge with fewer unproductive interactions.

We evaluate our methods on two complementary settings: challenging question answering on the 100-query BrowseComp-Plus setting following RISE~\cite{BrowseCompP,RISE}, including a controlled expansion from 100K to 1M documents, and reasoning-intensive retrieval on four BRIGHT domains~\cite{BRIGHT}. As shown in Fig.~\ref{fig:cost}, with GPT-5.4-mini on the 100K-document BrowseComp-Plus corpus, \methodpp~achieves 84\% accuracy, compared with 78\% for both RISE and DCI, while requiring 23.9 average tool calls compared with 28.7 and 99.1, respectively. With GPT-5.4, \methodpp~reaches 91\% accuracy, exceeding RISE by 9 points. After expanding the corpus to 1M documents, \methodpp~retains 79\% accuracy, compared with 69\% for RISE. On BRIGHT, \methodp~achieves 53.36 average nDCG@10, outperforming DCI, RISE, and the retrieval-specialized NeMo agent. The gains in accuracy and tool efficiency indicate that relevance-aware interaction enables both faster and more reliable search convergence, while also improving the identification of reasoning-intensive evidence.

Our contributions are threefold:
\begin{itemize}[leftmargin=*,nosep]
    \item We identify a missing role of relevance in agentic search: beyond selecting retrieved content or managing a workspace, relevance can directly guide the execution and observation of corpus interactions.
    \item We propose \method, a relevance-aware \rg~search agent that preserves document rankings during corpus traversal, and introduce entry-point initialization and match-level reranking to provide coarse-to-fine guidance for faster and more reliable search convergence.
    \item We demonstrate a stronger accuracy--efficiency trade-off than retrieval-only agents, unrestricted DCI, and retrieval-constructed interaction spaces across challenging QA, corpus scaling, and reasoning-intensive retrieval.
\end{itemize}

\section{Related Work}

\subsection{From Document Retrieval to Agentic Retrieval}

Relevance estimation is a central organizing principle of modern information retrieval. Sparse methods such as BM25 rank documents using lexical statistics~\citep{robertson2009probabilistic}, whereas dense retrievers encode queries and documents into a shared representation space for semantic matching~\citep{karpukhin2020dense}. 

Retrieval-augmented generation uses such retrievers to ground language models in external knowledge and has been studied in settings involving long inputs~\cite{yu2025prelude,ding2026exdr,chung2025divlogiceval}, many-shot demonstrations~\cite{chungmany}, context compression~\cite{chung2024selection}, and globally distributed evidence~\citep{lewis2020retrieval}.
However, these approaches still rely on a context assembled before reasoning that must anticipate downstream evidence needs, while longer ranked contexts do not guarantee that a model will reliably use decisive evidence buried among less useful material~\citep{liu2024lost}.

Agentic retrieval relaxes this one-shot assumption by allowing a model to control retrieval over multiple steps, interleaving reasoning with retrieval actions so that intermediate deductions can shape subsequent information needs~\citep{yao2022react,trivedi2023interleaving,zhou2025hgmem,zhang2026expseek,zeng2026rethinking}. Yet many such systems still treat a conventional retriever as the primary corpus interface, returning top-$k$ results over which the agent can reason only after retrieval. This design can conflate document-level relevance with evidence utility: a decisive clue may lie in a short span omitted during passage selection, while a document's value may become apparent only after an intermediate entity is discovered. These limitations motivate corpus interfaces that combine scalable corpus-level prioritization with finer-grained evidence localization, composition, and verification.

\subsection{Corpus Interfaces for Agentic Search}

Beyond retrieval quality, a key design axis is the \emph{resolution} of the interface through which an agent accesses the corpus. Retriever-mediated systems, including strong lexical pipelines such as Pi-Serini~\citep{PiSerini}, expose ranked document views and thus confine the agent largely to evidence selected in advance. Direct Corpus Interaction (DCI) replaces this constraint with a high-resolution interface, exposing the raw corpus as files that an agent manipulates through pattern matching, local reads, and shell commands, thereby supporting exact lexical constraints, entity following, and cross-file composition~\citep{DCI}. GrepSeek further shows that such interaction policies can be learned by a compact agent through verified trajectories and reinforcement learning~\citep{salemi2026grepseek}. This flexibility, however, comes with limited built-in corpus-level prioritization: unrestricted operations may traverse the corpus as if all files were equally promising, causing cost to grow with the number of distractors.

RISE~\citep{RISE} and DR-DCI~\cite{DRDCI} address this scalability tension by bounding direct interaction. RISE reframes retrieval as constructing an \emph{interaction space}, using BM25 to select a subset preprocessed for shell-style navigation, whereas DR-DCI treats retrieval as an agent-callable action that progressively expands a persistent workspace. In both, relevance is used primarily to construct or expand the searchable space, but is not explicitly propagated to order local traversal or prioritize matches when raw output exceeds the observation budget.

This exposes a gap between \emph{estimating} corpus-level relevance and \emph{using} it during interaction. Rather than treating relevance as a one-time filter on the searchable corpus, our approach carries it into direct corpus interaction as an execution prior, guiding where search begins, the order in which documents are traversed, and which local matches reach the model under a limited observation budget.


\begin{figure}
    \centering
    \includegraphics[width=1\linewidth]{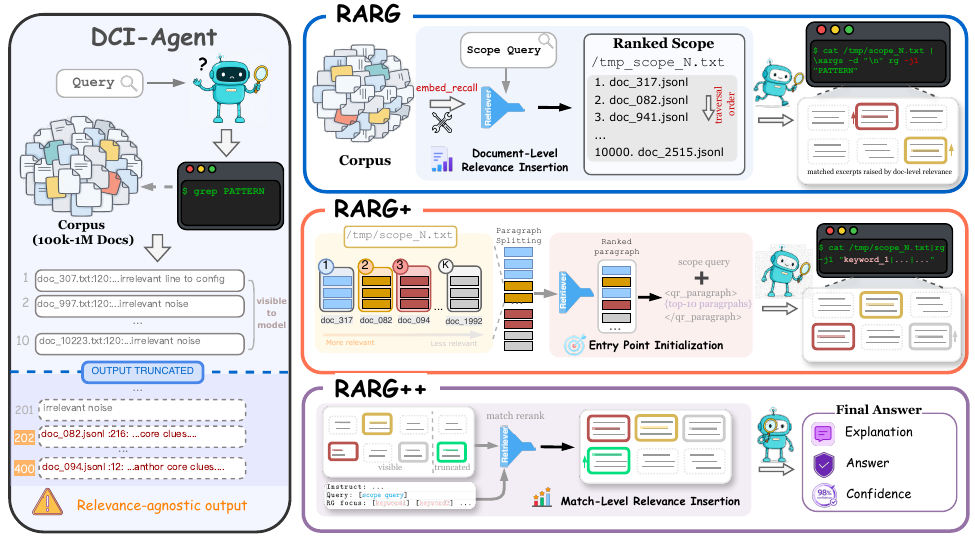}
    \caption{Overview of \method. \textit{embed\_recall} ranks the corpus into a scope file, and \rg~scans it in ranked order so relevant documents surface first (\method). \methodp~adds query-relevant paragraphs as an entry point; \methodpp~reranks \rg~matches to keep informative excerpts from lower-ranked documents visible. Document-level relevance sets where \rg~searches first; match-level relevance sets which matches reach the model.}
    \label{fig:framework}
\end{figure}

\section{Method}\label{sec:method}

Since \grep~matches keywords or patterns rather than relevance, its hits need not bear on the question, forcing the agent through idle search steps. We therefore introduce the Relevance-Aware RipGrep (RARG) agent as illustrated in Fig.~\ref{fig:framework}, which guides \grep~toward better and faster exploration.

\subsection{DCI-based Agent Structure}\label{subsec:dci_str}

RARG is built entirely on DCI-Agent-Lite, a search agent that exposes only two tools: \textit{Bash}(~\lstterm[black]{command}~) and \textit{Read}(~\lstterm[black]{doc\_path}, \lstterm[black]{offset}, \lstterm[black]{limit}~). \textit{Bash} mainly issues \grep~commands to find local matches across the corpus, and \textit{Read} returns lines~\lstterm[black]{offset}~to~\lstterm[black]{offset+limit}~of a document when a match calls for more context.

Since the tool arguments are generated freely by the LLM, \textit{Bash} may also receive commands such as $\{$\texttt{ls, find, python, cat, ...}$\}$. Because the corpus can be very large, running \texttt{ls}/\texttt{find} over the whole corpus can stall for a long time; we therefore forbid \texttt{ls}/\texttt{find} whenever they target the entire corpus.

The agent keeps issuing tool calls until it is confident enough to answer, but unbounded interaction quickly overflows the context. DCI-Agent-Lite mitigates this with four levels of context management; we adopt the default Level-3 compaction: (1) truncate a single tool result to $C$ characters; (2) once the context length exceeds a threshold, retain the tool results of the most recent $T$ turns and replace older ones with ``$[$cleared$]$'' placeholders.

\subsection{Document-Level Relevance Insertion}\label{subsec:doc_level}

As argued above, relevance remains essential for \grep~to locate clues faster and more reliably; the central question is how to inject relevance into \grep. Embedding retrieval, the most common way to rank documents by relevance to a question, is known to underperform plain \grep~when used to feed top-ranked documents directly to the LLM~\cite{DCI}. We instead treat retrieval as a ``support'' that supplies relevance guidance to \grep~(the ``carry''), rather than as the evidence channel itself.

We first inject document-level relevance into \grep~via a new tool, \textit{embed\_recall}(~\lstterm[black]{scope\_query}~), which takes an LLM-generated query. Rather than returning document contents, it writes the ranked document paths to a temporary file \texttt{$\backslash$tmp$\backslash$scope\_\{N\}.txt} and returns only the mapping ``\texttt{$\backslash$tmp$\backslash$scope\_\{N\}.txt} $\rightarrow$ query'' to the LLM. Recording all paths is unnecessary for very large corpora, both to limit processing cost and because relevant documents almost always rank near the top; we thus cap the scope at 10{,}000 paths, which still yields very high evidence recall---hence the term ``scope''. The LLM is then asked to \texttt{cat} the scope file and \textbf{search along the path order}, so that matches in more semantically similar documents surface earlier. Concretely, we instruct the LLM to use the following command:
\begin{center}
\begin{lstlisting}[language=bash]
(*\textcolor{blue}{cat /tmp/scope\_\{N\}.txt}*) | (*\textcolor{blue}{xargs -d `\textbackslash n'}*) (*\textcolor{orange}{rg [OPTIONS] "PATTERN"}*)
\end{lstlisting}
\end{center}
This appends every doc path in the scope to the \rg~block, restricting the search to those documents. However, \rg~reads files in parallel (multi-threaded) and emits matches in the order threads finish, which destroys the path ordering we inject. We therefore add a rule-based pass that regex-matches \rg~invocations and injects \lstterm[orange]{-j1} to force sequential, single-threaded scanning (\lstterm[orange]{rg -j1 [OPTIONS]}).

The LLM is instructed to begin each episode by calling \textit{embed\_recall} with the original question and then \rg-searching (with the internal \lstterm[orange]{-j1} insertion) within the resulting scope; it may call \textit{embed\_recall} again to form a new scope once \rg~exhausts the current angles. During compaction, the ``\texttt{$\backslash$tmp$\backslash$scope\_\{N\}.txt} $\rightarrow$ scope query'' mappings are retained rather than cleared, so the LLM always knows which scopes are available. We call this method \method. Refer to prompts and tool definitions in Appendix~\ref{app:prompt}.

\subsection{Entry Point Initialization}

The agent starts from the question alone, and \textit{embed\_recall} adds no content to enrich it, so \rg~may waste several steps before finding a good entry point for convergence. To supply such an entry point without an extra LLM turn, we append a few query-relevant paragraphs from the top-ranked scope documents to the \textit{embed\_recall} output.

Specifically, we split the top-$X$ documents of a scope into paragraphs of 400--1000 characters, encode them with the embedding model, score them against the scope query, and keep the top-10. These paragraphs are wrapped in <qr\_paragraph></qr\_paragraph> and appended after the scope mapping, serving as a starting point for the LLM rather than as the answer itself. They are cleared upon compaction. We denote \method~with this initialization as \methodp.

\subsection{Match-Level Relevance Insertion}

To further speed up convergence, each search step should also account for the fine-grained relevance between individual matched excerpts and the search goal. As noted in~\ref{subsec:dci_str}, the \rg~output is truncated and keeps only some of the matches, especially within the top-ranked documents of \method. Under document-level relevance alone, an informative excerpt can be diluted by the surrounding document content in the embedding space, giving its document a low rank and leaving the excerpt invisible. We therefore use the embedding model to rerank a broader pool of matches and present only the top ones.

A challenge of per-step match reranking is that the embedding model needs a dedicated query capturing both the global search goal and the local \rg~intent. We explore two options: (1) a constructed query and (2) a generative query.

By construction, an \textit{embed\_recall} call yields a scope and its query, which \rg~then explores over several steps; the scope query captures the global goal, and the \rg~keywords/patterns capture the local intent. Hence, when \rg~\texttt{cat}s a \texttt{$\backslash$tmp$\backslash$scope\_\{N\}.txt}, we convert its patterns into a keyword sequence by rules and concatenate them with the scope query as:
\begin{center}
\begin{lstlisting}[language=bash]
Instruct: ...
Query: [(*\textcolor{cyan}{scope query}*)]
RG focus: [(*\textcolor{pink}{keyword1}*)] [(*\textcolor{pink}{keyword2}*)] ...
\end{lstlisting}
\end{center}
This constructed query reranks up to $M$ matches and selects the top $m$ ($m<M$), giving more candidate clues a chance to reach the LLM. Reranking is skipped when the number of matches is below $m$.

For the generative query, we replace \textit{Bash} with \textit{RerankAwareBash}(~\lstterm[black]{command}, \lstterm[black]{rerank\_query}=None~), where the LLM supplies \lstterm[black]{rerank\_query} only when it intends to run \rg~and leaves it None otherwise. The LLM is instructed to write a \lstterm[black]{rerank\_query} that reflects both the final goal and the local retrieval intent. In our experiments, however, this generative variant degrades performance. We denote \method~with match-level reranking as \methodpp, using the constructed query by default.

\section{Experiments}

\subsection{Experiment Setup}

\textbf{Benchmarks.}~~~~We evaluate the effectiveness and efficiency of our methods in two scenarios: challenging browse question answering and reasoning-intensive retrieval. For QA, we follow RISE and use BrowseComp-Plus~\cite{BrowseCompP} (BC+), which answers difficult browsing questions over a fixed 100K-document corpus; due to cost, RISE evaluates on a 100-query sample. To study larger search spaces, we further expand the corpus with 900K long, noisy documents sampled from FineWeb-Edu, as longer documents make search harder by offering more opportunities for incidental matches against noisy content. We compute accuracy with the LLM-as-judge (GPT-5.1) prompt of DCI. For retrieval, we follow DCI and use four BRIGHT~\cite{BRIGHT} subsets: biology, earth science, economics, and robotics. BRIGHT is harder than conventional IR benchmarks, as it demands in-depth reasoning rather than shallow semantic matching. The four subsets contain 103/116/103/101 questions over corpora of 57K/121K/50K/62K documents, and we report nDCG@10.

\textbf{Implementation Details.}~~~~We run all experiments with GPT-5.4-mini (medium thinking effort), GPT-5.4-nano (high thinking effort)~\cite{openai2026gpt54nano}, and GPT-5.4 (medium thinking effort)~\cite{openai2026gpt54}. As described in Section~\ref{sec:method}, \method~is built on the DCI-Agent-Lite harness, with a few adaptations. For \rg, we cap the number of returned matches at 30 for BC+ and 60 for BRIGHT, and truncate each match to 1{,}000 characters for BC+ and 500 for BRIGHT; the resulting output is at most 30{,}000 + 30\,(60)$\times$path\_name\_length characters, far below DCI's 50\,KB. DCI compaction keeps the tool results of the most recent 12 turns, but since GPT often calls several tools per turn, this retains too many results; we instead keep the most recent 40 tool results and raise the compaction threshold to 230K to better exploit the cache. For BC+, we use Qwen3-Embedding-4B~\cite{Qwen3Emb} (Q3E) for document ranking, initialization, and match reranking. For BRIGHT, we use llama-nv-embed-reasoning-3b\footnote{https://huggingface.co/nvidia/llama-nv-embed-reasoning-3b} (NV) for document ranking and initialization, and Q3E for match reranking. NV is a reasoning-intensive embedding model used by the strongest BRIGHT agent, NeMo~\cite{NeMo}, whose setting we follow; we found NV poor at ranking short matches, hence the fallback to Q3E for reranking. We set the reranking pool to $M$=500 with $m$=30/60, and cap the number of turns at 100. \method~is implemented in Python and tested on Linux with one H20 GPU.


\textbf{Compared Agent Methods.}~~~~We compare against the following agents. \textbf{DCI-Agent-Lite}~\cite{DCI} is a pure \grep~agent equipped only with \textit{Bash} and \textit{Read}. \textbf{RISE}~\cite{RISE} builds an interaction space via BM25 retrieval and augments the retrieved documents with navigational structure, which the agent then explores with shell tools. \textbf{RISE-BM25} is the ablation that keeps the BM25 interaction space but uses the original, unstructured documents without this navigational processing. We run both DCI-Agent-Lite and the RISE-based baselines using the official RISE implementation~\footnote{\url{https://github.com/texttron/RISE}}. We additionally evaluate \textbf{RISE-Q3-Emb-4B}, which replaces the BM25 retriever with Q3E, and \textbf{Retrieval-Agent}, the conventional retrieval baseline released with RISE. \textbf{Embedding Agent} is an agent that retrieves candidate documents through Q3E-based dense retrieval. \textbf{NeMo Agent}~\cite{NeMo} is a retrieval-oriented agent that iteratively recalls and ranks candidate documents to form the final output; we use its official implementation~\footnote{\url{https://github.com/NVIDIA/NeMo-Retriever/tree/main/retrieval-bench}}. Implementation details of these agents are provided in Appendix~\ref{app:agent_impl}.

\begin{table*}[t]
  \centering
  \small
  \setlength{\tabcolsep}{4pt}
  \caption{Main results on 100-query (RISE version) BrowseComp-Plus with a 100K-document corpus. Turns and Tools denote the average agent turns and tool calls. Search, Bash, and Read report the average number of calls to each tool. For our methods, Search calls are \textit{embed\_recall}. The highest Acc and the fewest Turns/Tools are in bold; Parts of Search/Bash are underlined to show different behaviors of agents. $^{\triangle}$ denotes the results are cited from RISE~\cite{RISE}; $^{?}$ denotes that RISE does not mention what thinking effort is activated. The left tool calls of NeMo are: \textit{Think} - 1.3/0.5 (mini/nano) and \textit{Answer} - 0.8/0.7.}
  \label{tab:bc100k-main}
  \scalebox{0.8}{
  \begin{tabular}{c|l|ccc|ccc}
    \toprule
    Model & Method & \textbf{Acc}$\uparrow$ & Turns & Tools & Search & Bash & Read \\
    \midrule
    \multirow{8}{*}{\shortstack{GPT-5.4 mini\\(medium effort)}} 
    & RISE$^{\triangle}$               & 78 & 24.3 & 28.7 & 13.1 &  9.2 & 6.4 \\
    & RISE-BM25$^{\triangle}$          & 77 & 23.0 & 29.6 & \underline{14.8} &  \underline{9.6} & 5.2 \\
    & RISE-Q3-Emb-4B                   & 69 & 28.9 & 35.9 & 22.1 &  9.1 & 4.7 \\
    & Retrieval-Agent$^{\triangle}$    & 68 & 29.2 & 38.9 & 37.0 &   -- & 1.9 \\
    & Embedding Agent                  & 55 & 54.3 & 68.0 & 68.0 & --   & --  \\
    & Nemo Agent                       & 64 & 25.3 & 30.3 & 28.2 & --   &  -- \\
    & DCI$^{\triangle}$                & 78 & 48.8 & 99.1 &   -- & \underline{90.3} & 8.8 \\
    & \cellcolor{gray!20}\method    & \cellcolor{gray!20}80 & \cellcolor{gray!20}18.2 & \cellcolor{gray!20}29.8 & \cellcolor{gray!20}\underline{1.2} & \cellcolor{gray!20}\underline{27.2} & \cellcolor{gray!20}1.4 \\
    & \cellcolor{gray!20}\methodp   & \cellcolor{gray!20}81 & \cellcolor{gray!20}20.2 & \cellcolor{gray!20}29.6 & \cellcolor{gray!20}\underline{1.6} & \cellcolor{gray!20}\underline{26.6} & \cellcolor{gray!20}1.3 \\
    & \cellcolor{gray!20}\methodpp  & \cellcolor{gray!20}\textbf{84} & \cellcolor{gray!20}\textbf{17.6} & \cellcolor{gray!20}\textbf{23.9} & \cellcolor{gray!20}\underline{1.5} & \cellcolor{gray!20}\underline{21.1} & \cellcolor{gray!20}1.3 \\
    & \methodpp~(generative) & 75 & 15.8 & 17.8 & 1.2 & 15.4 & 1.3 \\
    \midrule
    \multirow{6}{*}{\shortstack{GPT-5.4-nano\\(high effort)}}
      & RISE$^{\triangle}$      & 68 & 23.4 &  28.7 & 10.4 &   8.4 & 4.6 \\
      & RISE-BM25$^{\triangle}$ & 64 & \textbf{22.0} &  \textbf{29.6} & 10.2 &   8.1 & 3.7 \\
      & Embedding Agent         & 49 & 43.1 & 66.6 & 66.6 & -- & -- \\
      & NeMo Agent              & 57 & 12.1 & 37.8 & 36.6 & -- & -- \\
      & DCI$^{\triangle}$       & 71 & 45.7 & 126.5 &   -- & 119.4 & 7.1 \\
      & \cellcolor{gray!20}\method       & \cellcolor{gray!20}73 & \cellcolor{gray!20}39.9 &  \cellcolor{gray!20}41.4 &  \cellcolor{gray!20}\underline{9.1} &  \cellcolor{gray!20}29.5 & \cellcolor{gray!20}2.8 \\
      & \cellcolor{gray!20}\methodp      & \cellcolor{gray!20}74 & \cellcolor{gray!20}40.2 &  \cellcolor{gray!20}39.8 & \cellcolor{gray!20}\underline{14.5} &  \cellcolor{gray!20}22.4 & \cellcolor{gray!20}2.9 \\
      & \cellcolor{gray!20}\methodpp     & \cellcolor{gray!20}\textbf{79} & \cellcolor{gray!20}36.0 & \cellcolor{gray!20}36.1 & \cellcolor{gray!20}\underline{9.6} & \cellcolor{gray!20}23.4 & \cellcolor{gray!20}3.1 \\
    \midrule
    \multirow{2}{*}{\shortstack{GPT-5.4\\(medium effort)}}
      & RISE$^{?\triangle}$  & 82 & 32.20 &  34.30 & \underline{11.00} &  \underline{16.20} & 7.10 \\
      & \cellcolor{gray!20}RARG++        & \cellcolor{gray!20}\textbf{91} & \cellcolor{gray!20}\textbf{13.59} & \cellcolor{gray!20}\textbf{25.43} & \cellcolor{gray!20}\underline{1.58} & \cellcolor{gray!20}\underline{19.46} & \cellcolor{gray!20}4.39 \\
    \bottomrule
  \end{tabular}%
  }
\end{table*}

\subsection{Results on Challenging QA}

Table~\ref{tab:bc100k-main} shows that RARG attains the best accuracy on every backbone while sharply reducing interaction cost. \methodpp~reaches 84\%/79\% on GPT-5.4-mini/nano and 91\% on GPT-5.4, exceeding the strongest baseline by 6/8/9 points, yet uses 23.9/25.43 tool calls against 99.1 for DCI and 28.7/34.3 for RISE (except the GPT-5.4-nano setting). This confirms our central claim: guiding interaction with relevance, rather than replacing it, converts the accuracy of relevance-agnostic DCI into a far cheaper search.

The Embedding Agent and NeMo Agent isolate the retrieval-centric interface. We include NeMo because its recall-and-rank harness deduplicates and recalls more distinct documents, letting us test whether such a stronger-coverage variant of the Embedding Agent models the evidence better. Yet both remain weakest on BC+: they expose relevance only as top-ranked content rather than a guide for exploration, exactly the interface bottleneck claimed by \citet{DCI}.

The underlined \textit{Search}/\textit{Bash} counts expose \emph{how} relevance is used. RISE uses retrieval to build a working space that bounds the corpus and reduces \rg~exploration steps. Yet every space-constructing \textit{Search} also returns snippets (the first few hundred characters) of the top documents, so retrieval simultaneously serves as an information channel to the model. This has the risk of incentivizing issuing more \textit{Search} calls, since each one directly supplies content. As we use Q3E while RISE was designed around BM25, we further run RISE-Q3-Emb-4B to control for retriever strength; it does not help and in fact issues more \textit{Search} calls, indicating that RISE stays adapted to its BM25 interface rather than a stronger dense retriever. RARG inverts this balance: except on nano, it issues only 1.2--1.6 \textit{embed\_recall} calls to fix a document-level search order, then delegates the bulk of exploration to scoped \rg. Relevance thus acts as an execution prior over \rg~traversal, not as the evidence channel itself. On nano, the pattern is less clean (higher and more variable \textit{embed\_recall}), which we attribute to weaker instruction following and revisit in the behavior analysis. Despite this, our methods with GPT-5.4-nano still achieve the best performance.

Within RARG, the variants trace the intended coarse-to-fine progression: from \method~to \methodp~to \methodpp, mini accuracy rises 80$\rightarrow$81$\rightarrow$84 while tools fall 29.8$\rightarrow$29.6$\rightarrow$23.9, and nano shows the same ordering (73/74/79). Each level of relevance therefore adds accuracy \emph{and} accelerates convergence, directly echoing our three contributions. The generative variant breaks this trend: it converges fastest (17.8 tools) but drops to 75\%. We attribute this to a train--evaluation gap---forcing the model to emit an explicit rerank query perturbs the \textit{Bash}/\rg~behavior it was trained on. Since it does improve convergence, closing this gap is a promising direction for future work.

\subsection{Scaling the Corpus}\label{subsec:scaling}

To test whether the relevance-guided interaction remains useful as the search space grows, we expand the BC+ corpus from 100K to 1M documents. The additional 900K documents are long FineWeb-Edu articles, and therefore introduce substantially more opportunities for incidental lexical matches and irrelevant context.

\begin{table}[t]
  \centering
  \small
  \setlength{\tabcolsep}{4pt}
  \caption{Results after scaling 100-query BrowseComp-Plus from 100K to 1M documents
  using GPT-5.4-mini (medium reasoning).}
  \label{tab:bc1m}
  \vspace{0.2cm}
  \scalebox{0.8}{
  \begin{tabular}{c|l|ccc|ccc}
    \toprule
    Model & Method & \textbf{Acc}$\uparrow$ & Turns$\downarrow$ & Tools$\downarrow$ & Search & Bash & Read \\
    \midrule
    \multirow{4}{*}{\shortstack{GPT-5.4 mini\\(medium effort)}} 
    & RISE-BM25 & 69 & 25.4 & 32.1 (+2.5) & 20.3 (+5.5) &  8.3 (-1.3) & 3.5 (-1.7) \\
    & \cellcolor{gray!20}\method
      & \cellcolor{gray!20}78 & \cellcolor{gray!20}19.3 & \cellcolor{gray!20}31.8 (+2.0) & \cellcolor{gray!20}1.5 (+0.3) & \cellcolor{gray!20}28.7 (+1.5) & \cellcolor{gray!20}1.7 (+0.3) \\
    & \cellcolor{gray!20}\methodp
      & \cellcolor{gray!20}78 & \cellcolor{gray!20}21.9 & \cellcolor{gray!20}30.4 (+0.8) &  \cellcolor{gray!20}1.5 (-0.1) & \cellcolor{gray!20}27.5 (+0.9) & \cellcolor{gray!20}1.4 (+0.1) \\
    & \cellcolor{gray!20}\methodpp
      & \cellcolor{gray!20}\textbf{79} & \cellcolor{gray!20}\textbf{17.8} & \cellcolor{gray!20}\textbf{24.7} (+0.8)
      & \cellcolor{gray!20}1.4 (-0.1) & \cellcolor{gray!20}22.0 (+0.9) & \cellcolor{gray!20}1.4 (+0.1) \\
    \bottomrule
  \end{tabular}
  }
\end{table}

As shown in Table~\ref{tab:bc1m}, scaling to 1M documents lowers all methods, with RISE-BM25 falling from 77\% to 69\% and \method/\methodp/\methodpp~from 80/81/84 to 78/78/79. This contradicts the non-decreasing trend RISE~\cite{RISE} reports under scaling. The difference may come from the design: our 900K added documents are \emph{long} FineWeb-Edu articles, which greatly increase the chance of incidental lexical matches and inject far more distracting context into each \rg~scan. Although performance drops for all search agents, the cost seems to be less affected by scaling, because the number of tools increases only slightly. We think the reason may be that retrievers (Q3E or BM25) are less affected, but \rg~is not immune to distractors---more noise per file degrades match quality just as it would any local lexical search. The distracted local matches introduce error to delude the judgment of LLMs. We analyze RARG's behavior on the 1M corpus in Section~\ref{subsec:analysis} to support the claim. Although \methodpp degrades with corpus scaling, it keeps a 10-point margin over RISE-BM25. Corpus-induced interference is only partially absorbed by match-level reranking, and therefore, how to reduce this weakness of local matching can be future work to study.

\subsection{Results on Reason-Intensive Retrieval}

BRIGHT and BC+ reward opposite search shapes. In ranking-style retrieval, the agent is encouraged to first recall as many relevant documents as possible and then order them precisely, which can be seen as a ``breadth-first'' process. In contrast, QA is more ``depth-first'': a single verified and concise reasoning path to reach the right answer suffices. Fast convergence, which RARG optimizes for, is therefore less aligned with BRIGHT, yet Table~\ref{tab:bright-ndcg} shows \methodp~still attains the best average nDCG@10 (53.36), ahead of the retrieval-specialized NeMo (52.89) and all other baselines. Relevance-aware interaction thus remains beneficial even when the objective is a ranked list rather than one answer.

Tool counts here reflect different interfaces, not comparable costs. NeMo is a recall-and-rank agent: it retrieves 20 full documents per \textit{Search} and leans on strong LLM reranking, and because BRIGHT documents are relatively short (unlike BC+), this loop is highly compact (7.68 tools). A grep-oriented agent instead observes only local matches and must \textit{Read} for surrounding context, hence more \textit{Read} tool calls. For DCI, the relevance-agnostic \rg~shows a steadily increasing number of exploration steps but provides less relevant information, resulting in lower performance. RISE still relies on \textit{Search}, but BM25 clearly cannot handle the reason-intensive retrieval. Within RARG, the ordering flips relative to QA: \methodp~$>$~\method~$>$~\methodpp~(53.36/51.75/50.55). \methodp, provided with a good initialization, achieves the best performance. However, we can see from the number of \textit{Bash} calls that \methodpp's match-level reranking concentrates the observation budget on locally strong excerpts and converges fastest relatively---an advantage for QA but a liability when the task rewards wide recall. Overall, RARG leverages relevance to top BRIGHT, but the optimal granularity depends on whether the task favors rapid convergence or broad candidate recall.

\begin{table*}[t]
  \centering
  \small
  \setlength{\tabcolsep}{6pt}
  \caption{Retrieval effectiveness on BRIGHT, measured by NDCG@10.
  Unless specified otherwise, methods use GPT-5.4-mini (medium reasoning).}
  \label{tab:bright-ndcg}
  \scalebox{0.8}{
  \begin{tabular}{l|c|cccc|c|ccccc}
    \toprule
    Method & \textbf{Avg.}$\uparrow$ & Bio. & Earth & Eco. & Rob. & Tools & Search & Bash & Read & Think & Answer \\
    \midrule
    DCI & 48.43 & 62.05 & 54.94 & 37.13 & 39.59 & 40.04 & -- & 14.73 & 25.31 & -- & -- \\
    RISE-BM25 & 41.60 & 50.27 & 47.80 & 33.65 & 34.67 & 31.82 & 7.55 & 2.15 & 22.12 & -- & -- \\
    Nemo Agent & 52.89 & 65.15 & 61.85 & \textbf{39.05} & 45.49 & 7.68 & 6.36 & -- & -- & 0.37 & 0.95 \\
    \midrule
    \rowcolor{gray!20}\method & 51.75 & 63.87 & 60.54 & 38.50 & 44.07 & 28.73 & 1.10 & 11.25 & 16.38 & -- & -- \\
    \rowcolor{gray!20}\methodp & \textbf{53.36} & \textbf{66.70} & \textbf{62.16} & 37.23 & \textbf{47.34} & 27.55 & 1.23 & 9.87 & 16.45 & -- & -- \\
    \rowcolor{gray!20}\methodpp & 50.55 & 61.65 & 61.32 & 36.14 & 43.10 & 27.28 & 1.14 & 8.40 & 17.74 & -- & -- \\
    \bottomrule
  \end{tabular}
  }
\end{table*}

\subsection{Analysis of \method~Behavior}\label{subsec:analysis}

\begin{figure}
    \centering
    \includegraphics[width=1\linewidth]{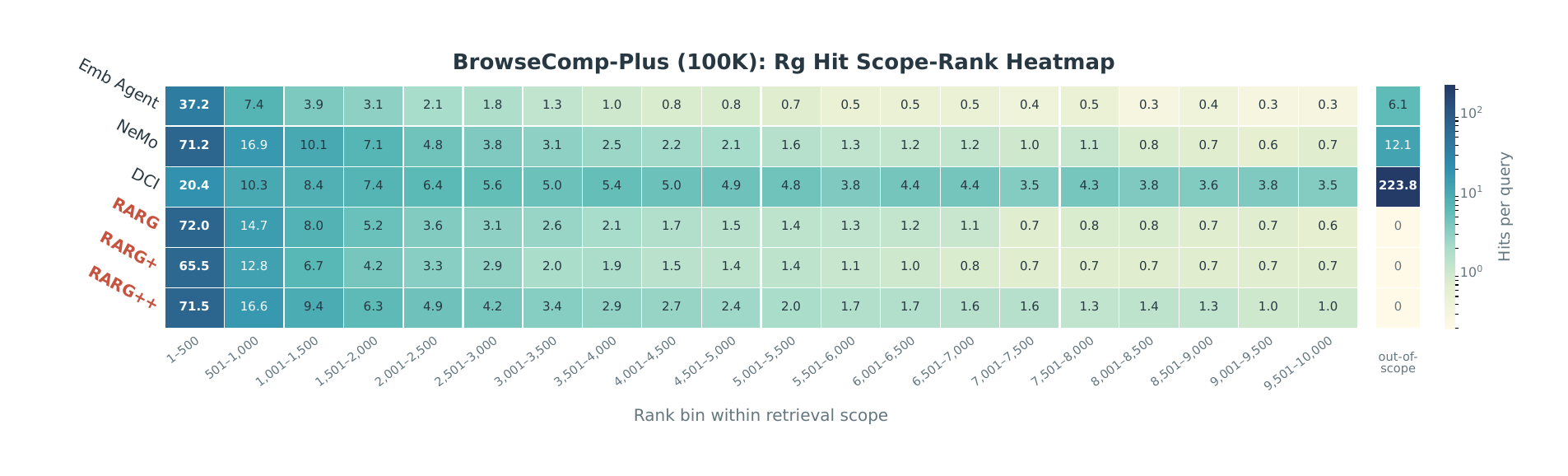}
    \caption{The updated relevant-document hits over scope ranks on BC+. Non-scope methods use the \texttt{scope\_1.txt} of \method. Embedding hits are top-concentrated but few, DCI is flat and relevance-agnostic, and RARG stays front-loaded while match-level reranking recovers evidence from lower-ranked documents.}
    \label{fig:scope_distribution}
\end{figure}

\textbf{How relevance shapes evidence discovery.}~~
To characterize \emph{how} each agent uses relevance, we bucket the scope documents and count where an agent's hits on relevant documents fall along the scope rank (Fig.~\ref{fig:scope_distribution}). The embedding agent, which consumes relevance directly, produces a sharply decreasing hit profile as expected; however, its distinct hits are few, indicating that its top-$k$ recall returns highly redundant documents and thus wastes the observation on repeated information. NeMo's deduplicated recall-and-rank harness does grant access to more distinct documents, but it remains bound by the same relevance-as-content interface of the emb agent. DCI shows the flattest profile and hits many out-of-scope documents, the signature of relevance-agnostic \grep~exploration: lacking any ordering prior, it must probe the whole corpus, yielding sparse evidence density and slow convergence. RARG sits between these extremes. It keeps hits front-loaded, confirming that document-level relevance steers traversal, while covering more distinct documents than the embedding agent. \methodpp~is the flattest of the three variants and recovers hits from lower-ranked documents: match-level reranking lets locally informative excerpts, buried in documents whose \emph{document}-level score is weak, be reranked forward and reach the model. This directly validates our design---global relevance orders traversal, local relevance governs visibility---preserving DCI's fine-grained interaction without discarding the relevance prior.

\begin{figure}
    \centering
    \includegraphics[width=1\linewidth]{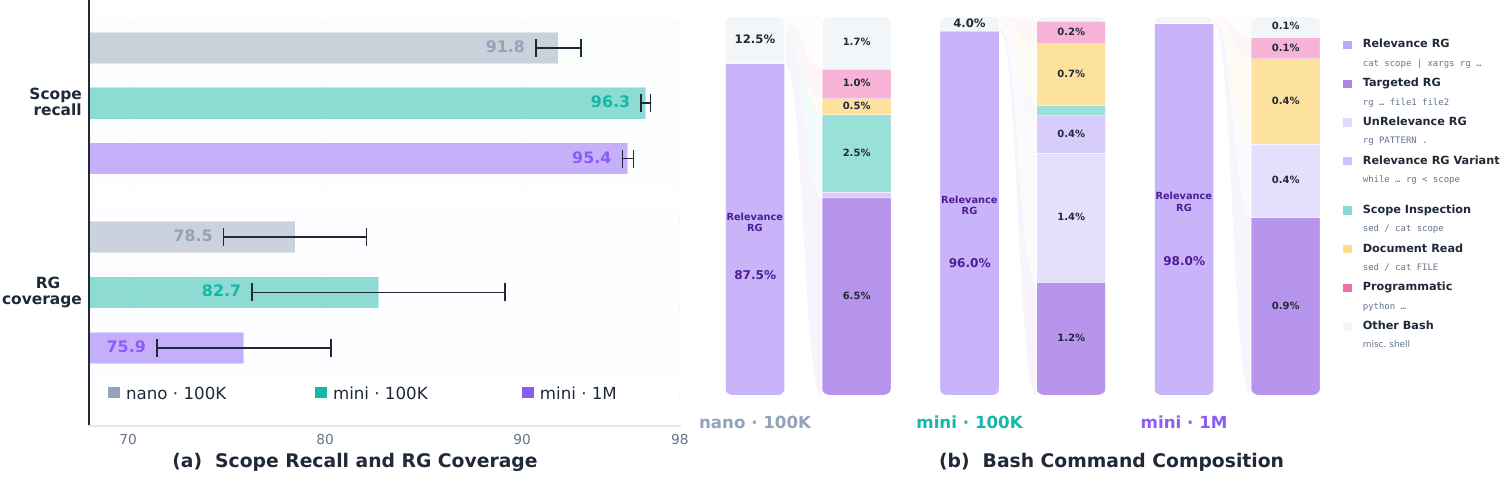}
    \caption{Scope quality and Bash usage under RARG on BrowseComp-Plus. \textbf{(a)}~Scope recall and RG coverage for GPT-5.4-nano and GPT-5.4-mini at the 100K corpus, and for mini at 1M. Error bars denote the standard deviation across RARG, RARG+, and RARG++. \textbf{(b)}~Composition of Bash calls: share of scoped \texttt{rg} versus
non-scoped Bash, with the residual mix broken down by command type.
Larger backbones and corpora keep scoped \texttt{rg} dominant.}
    \label{fig:portion}
\end{figure}

\textbf{Scope quality and Bash composition.}~~~
Fig.~\ref{fig:portion}(a) reports two per-scope quantities, averaged within each sample and then across samples: \emph{scope recall}, the fraction of BC+ gold documents captured by the scope, and \emph{RG coverage}, the fraction recovered by the files that \rg~actually hits. Both mini settings reach 95--97\% scope recall, showing that capping the scope at the top-10K documents preserves near-complete gold coverage while keeping the \rg~search load not too heavy; the negligible 100K$\rightarrow$1M drop further confirms that the retriever is barely perturbed by the added distractors. Nano recall is markedly lower, consistent with weaker \textit{embed\_recall} queries and its frequent, less disciplined \textit{embed\_recall} triggering. RG coverage tells a complementary story: mini-100K stays high, but mini-1M falls sharply---the long FineWeb-Edu documents flood \rg~with incidental matches, so the agent recovers fewer gold files per scan and may prematurely conclude that no further evidence exists, which explains the claim we stated in Sec.~\ref{subsec:scaling}. Fig.~\ref{fig:portion}(b) decomposes \textit{Bash} usage. On both mini settings, the harness reliably steers the model into the relevance-aware form \texttt{cat scope.txt | xargs -d '$\backslash$n' rg ...}: scoped \rg~dominates; the remaining non-scoped calls are still mostly \rg. On nano, the scoped-\rg~share shrinks substantially, again reflecting weaker instruction adherence rather than a limitation of the mechanism. Together, the two panels confirm that RARG supplies high-recall, low-cost scopes and converts them into relevance-ordered \rg~exploration whenever the backbone follows instructions well.
\subsection{Case Study}
We qualitatively compare the three variants on a 100K-document BC+ question that identifies a mathematician by joining an AMS fellowship and Ph.D.\ year with two coauthors, an award, and a second paper title. All three trajectories are judged correct for Russell David Lyons, while tool use decreases strictly from 33 to 18 to 10 and total turns fall from 17 to 11 to 10. The answer-bearing CV first appears at turn 7 for \method, but at turn 2 for both initialized variants; subsequent interaction verifies the Peres--Pemantle coauthor chain.

Appendix~\ref{app:case-study} presents the compressed trajectories, decisive observations, and interaction counts. The case is intended as a qualitative view of how progressively finer relevance guidance can shape search convergence, rather than as standalone evidence of method superiority; the aggregate comparison is reported in Table~\ref{tab:bc100k-main}.

\section{Conclusion}

We revisited the role of relevance in agentic search and argued that it should guide corpus interaction, not merely select its inputs. Building on this view, we introduced \method, which turns retrieval scores into an execution prior for \grep~exploration at two resolutions: document-level relevance orders \rg~traversal so promising documents are scanned first, while match-level reranking controls which local excerpts remain visible under a limited observation budget. Between them, entry-point initialization gives the agent a precise place to begin. This coarse-to-fine design preserves DCI's fine-grained, compositional interaction while restoring the relevance guidance that unrestricted \grep~lacks. Across challenging QA, corpus scaling, and reasoning-intensive retrieval, \method~advances the accuracy--efficiency frontier over retrieval-only agents, unrestricted DCI, and retrieval-constructed interaction spaces, reaching higher accuracy with substantially fewer tool calls and degrading gracefully as the corpus grows. Our analysis further confirms the intended mechanism: \method~keeps hits front-loaded toward relevant documents yet recovers useful evidence from lower-ranked ones. We hope these results encourage viewing relevance as an execution signal for corpus interaction, rather than a fixed content bottleneck.

\section*{Limitations}

\method's effectiveness depends on the quality of the relevance signal from the embedding model, and the two guidance levels place \emph{different} demands on it: document-level ranking must score long documents, whereas match-level reranking must score short local excerpts. These granularities are not always served equally well by the same model---on BRIGHT, for instance, NV ranks documents well but handles short matches less reliably, prompting a fallback to Qwen3-Embedding. Like other embedding-agent methods, \method~thus relies on a sufficiently strong embedding model, together with a harness adapted to it.

Enforcing single-threaded, order-preserving \rg~(\lstterm[orange]{-j1}) and reranking matches also incur extra search latency. The overhead is tolerable in our experiments, but a performance--time trade-off remains. We use \lstterm[orange]{-j1} rather than reordering matches after an unconstrained \rg~run because a query with many matches would make the full scan itself slow; sequential scanning instead lets us stop as soon as 30/60 (or $M$=500) matches are collected, achieving the same ordering with far less post-processing. Both are valid implementations, and other realizations are possible---the central idea of injecting relevance into \rg~is unchanged.

The method is further sensitive to the backbone's instruction-following ability. We find GPT-5.4-nano follows our multi-stage protocol less reliably than GPT-5.4-mini and GPT-5.4, which also converge faster; a more compliant model therefore yields cleaner behavior, though \method~still improves over baselines even on nano.

Relevance guidance is also not immune to corpus-induced interference. When the corpus is scaled with long, noisy documents, incidental lexical matches degrade both \rg~output and the embedding scores, and match-level reranking only partially mitigates this. Relatedly, the generative reranking-query variant converges fastest yet loses accuracy, suggesting a train--evaluation gap in which explicitly emitting a rerank query perturbs the learned \textit{Bash}/\rg~behavior; closing this gap is left to future work. Finally, our evaluation covers BrowseComp-Plus and four BRIGHT subsets with the GPT-5.x family; broader validation across other backbones, open-web settings, and more domains remains future work.

\bibliography{anthology}
\bibliographystyle{acl_natbib}

\appendix

\section{Agent Implementation Details}
\label{app:agent_impl}

We detail how the compared agents are configured for a fair comparison on both benchmarks.

\paragraph{BRIGHT prompting.} On BRIGHT, we found that the default DCI IR prompt performed substantially worse for the \grep-style agents. We therefore modified it to explicitly require the agent to rank 10 document names, matching the metric (nDCG@10) and the output format used by all methods.

\paragraph{Embedding Agent.} On BC+, we implement the Embedding Agent as a retrieval baseline that uses Qwen3-Embedding-4B (Q3E) as its retriever, without \grep-style corpus interaction. Each \textit{Search} call recalls 5 documents; we truncate each document by encoding it with the Qwen3 tokenizer, keeping the first 512 tokens, and decoding them back to text (the way that BC+~\cite{BrowseCompP} does). Following DCI/RARG, tool results of old turns are cleared once the context exceeds the compaction threshold, and the maximum number of turns is 100. The retrieved documents are \emph{not} deduplicated across \textit{Search} calls, so the recalled set carries a fairly high repetition rate, which wastes the observation budget on redundant content.

\paragraph{NeMo Agent.} We use the official NeMo implementation, with Q3E on BC+ and NV on BRIGHT, matching the embedding model each of our methods uses on the corresponding benchmark so that differences reflect the agent design rather than the retriever. Recalled documents are truncated in the same way as the Embedding Agent, i.e., keeping the first 512 Qwen3 tokens. As an agent specialized for retrieval, NeMo favors recalling more documents to raise recall and then reranking them to improve precision. Unlike the Embedding Agent, it deduplicates recalled documents: a repeated document is marked as already seen and its content is not re-shown, and NeMo keeps traversing until 5 \emph{new} documents are found each round. It therefore surfaces more distinct documents than the Embedding Agent.

We were curious how such a recall-and-rank agent behaves on a QA-style task like BC+, for which NeMo has no dedicated configuration; we therefore adapt it. The main issue is context overflow. NeMo's default ranking falls back to reciprocal rank fusion (RRF) and, failing that, invokes an extra selection agent. On BC+ we likewise use RRF: when the context is exceeded, we RRF-rank all recalled documents and feed them to a fresh answer agent; if overflow persists, we drop the lowest 5 documents in RRF order iteratively until they fit, and then answer. We also found it necessary to fix the number of returned documents to 5 per call---letting the model control the top-$k$ itself degrades performance.

\section{Prompts for \method}
\label{app:prompt}

\subsection{System Prompt of BC+}

Figure~\ref{fig:bc_prompt} shows the system prompt used by \method~on BrowseComp-Plus. It states the corpus-listing constraint, describes the \textit{embed\_recall} tool and the scope-based \rg~search protocol, and fixes the final answer format.

\begin{tcolorbox}[
  colback=gray!5, colframe=gray!55!black,
  boxrule=0.4pt, arc=2pt, left=4pt, right=4pt, top=3pt, bottom=3pt,
  breakable, fonttitle=\bfseries, title={System prompt for \method~on BrowseComp-Plus}]
\begin{lstlisting}[basicstyle=\ttfamily\scriptsize, escapeinside={}{}, backgroundcolor=\color{white}, frame=none, breaklines=true, columns=fullflexible]
IMPORTANT CONSTRAINTS:
- Do NOT use `ls`, `find`, or any command that lists all filenames. The corpus has 100,000+ files across 30,000+ directories -- listing them will timeout and waste all your turns.

EMBEDDING RECALL TOOL:
The corpus contains a large number of documents. The large search space puts significant pressure on rg -- too many matches, excessive noise, and difficulty pinpointing relevant files.

To address this, an embed_recall tool is available. By providing a query, it semantically recalls a batch of relevant documents and saves their paths to a scope file (scope_1.txt, scope_2.txt, ...), one path per line. This narrows the search space to a manageable subset. It also returns a few reference paragraphs from top-ranked documents -- these are a rough preview to help you identify search directions for `rg` search.

To search within recalled documents, pipe the scope file to rg:
  cat /path/to/scope_N.txt | xargs -d '\n' rg [OPTIONS] "PATTERN"

Results are returned ordered by document embedding similarity to your query (most relevant first). Only search within one scope file at a time -- do not concatenate multiple scope files.

NOTE:
- Initially call embed_recall with the ORIGINAL QUESTION to establish a scope file, then use rg to explore within it with diverse keywords, entities, and patterns.
- `rg` calls from different angles are essential.
- Only call embed_recall again if your scope is clearly missing the topic entirely -- do NOT call it repeatedly hoping for better preview paragraphs, use rg instead.
- Do NOT call embed_recall multiple times without rg exploration in between.
- Do NOT run rg directly on the corpus without a scope file.

When you are confident to answer the question, stop tool calling and respond with the following format:
Explanation: {your explanation for your final answer}
Exact Answer: {your succinct, final answer}
Confidence: {confidence score between 0% and 100%}
\end{lstlisting}
\end{tcolorbox}
\captionof{figure}{The system prompt used by \method~for the BrowseComp-Plus evaluation.}
\label{fig:bc_prompt}

\subsection{System Prompt of BRIGHT}

Figure~\ref{fig:bright_prompt} shows the system prompt used by \method~on BRIGHT. Unlike the QA prompt, it frames the task as retrieval: the agent must maximize recall of related documents while preserving precision, follows the same \textit{embed\_recall} plus scope-based \rg~protocol, and returns a ranked list of exactly ten document paths.

\begin{tcolorbox}[
  colback=gray!5, colframe=gray!55!black,
  boxrule=0.4pt, arc=2pt, left=4pt, right=4pt, top=3pt, bottom=3pt,
  breakable, fonttitle=\bfseries, title={System prompt for \method~on BRIGHT}]
\begin{lstlisting}[basicstyle=\ttfamily\scriptsize, escapeinside={}{}, backgroundcolor=\color{white}, frame=none, breaklines=true, columns=fullflexible]
You are a retrieval agent that finds all documents related to a given query.

<GOAL>
- You are given a search query. Your task is to use the available tools to find all related and somewhat related documents for that query while producing a strong final ranking.
- The goal is to maximize recall without throwing away precision: missing relevant documents hurts, and including weakly related documents also hurts.
</GOAL>

IMPORTANT CONSTRAINTS:
- Do NOT use `ls`, `find`, or any command that lists all filenames. The corpus has 100,000+ files across 30,000+ directories -- listing them will timeout and waste all your turns.

EMBEDDING RECALL TOOL:
The corpus contains a large number of documents. The large search space puts significant pressure on rg -- too many matches, excessive noise, and difficulty pinpointing relevant files.

To address this, an embed_recall tool is available. By providing a query, it semantically recalls a batch of relevant documents and saves their paths to a scope file (scope_1.txt, scope_2.txt, ...), one path per line. This narrows the search space to a manageable subset. It also returns a few reference paragraphs from top-ranked documents -- these are a rough preview to help you identify search directions for `rg` search.

To search within recalled documents, pipe the scope file to rg:
  cat /path/to/scope_N.txt | xargs -d '\n' rg [OPTIONS] "PATTERN"

Results are returned ordered by document embedding similarity to your query (most relevant first). Only search within one scope file at a time -- do not concatenate multiple scope files.

NOTE:
- Initially call embed_recall with the ORIGINAL QUERY to establish a scope file.
- Then use rg calls to explore within it with diverse keywords, entities, terminology, clues, paraphrases, aliases, and alternate phrasings.
- `rg` is the main search method.
- `rg` calls from different angles are essential.
- Only call embed_recall again if the current scope is clearly missing the wanted topic entirely. Do NOT call embed_recall repeatedly without rg exploration in between.
- Do NOT call embed_recall multiple times without rg exploration in between.
- Do NOT run rg directly on the corpus without a scope file.
- If needed, read the corresponding documents for more detailed evidences to ensure your ranking.

<RELEVANCE_DEFINITION>
- In this task, what counts as a "query", "document", and "relevant" can be more nuanced than in ordinary web search.
- A query may be a forum post, technical problem, math problem, coding problem, scientific question, or task description.
- A document is relevant if it is genuinely useful for answering, solving, grounding, or strongly supporting the query.
- Relevance is not limited to surface wording overlap. A document can be relevant because it provides the right entity, mechanism, theorem, example, fact, API, or other key clue needed for the query.
- You should analyze what usefulness means for this query and search for documents from multiple angles accordingly.
</RELEVANCE_DEFINITION>

<BEST_PRACTICES>
- You should be thorough and find all related and somewhat related documents.
- Search from different angles: direct entity names, aliases, abbreviations, alternate spellings, dates, technical terminology, paraphrases, and evidence chains.
- Do not stop after finding a few plausible documents if other relevant documents are likely to exist.
- At the same time, do not include tangential or weakly related documents just to make the list longer.
</BEST_PRACTICES>

<RANKING_INSTRUCTIONS>
- Rank the final list in decreasing order of relevance to the query.
- The first document should be the most useful one for the query, the second should be the next most useful one, and so on.
- All returned document identifiers must be full relative paths from the working directory and must correspond to actual corpus files.
- Return exactly the top 10 documents, ranked from most relevant to least relevant among your final selected set.
</RANKING_INSTRUCTIONS>

Your final response MUST follow this exact format:

Relevant Documents (ranked by relevance, most relevant first; exactly 10):
1. relative/path/to/doc1.txt
2. relative/path/to/doc2.txt
...
10. relative/path/to/doc10.txt
\end{lstlisting}
\end{tcolorbox}
\captionof{figure}{The system prompt used by \method~for the BRIGHT retrieval evaluation.}
\label{fig:bright_prompt}

\subsection{Tool Specifications}

We list the tool schemas exposed to the agent. \textit{embed\_recall} performs document-level relevance recall and writes a scope file; \textit{read} returns line-bounded file contents; and \textit{bash} executes shell commands, mainly \rg. The standard \textit{bash} is used by \method, \methodp, and \methodpp, while the \textit{RerankAwareBash} variant adds an optional \texttt{rerank\_query} for the generative match-level reranking of \methodpp.

\newtcolorbox{toolbox}[1]{
  colback=gray!5, colframe=gray!55!black,
  boxrule=0.4pt, arc=2pt, left=4pt, right=4pt, top=3pt, bottom=3pt,
  breakable, fonttitle=\bfseries, title={#1}}

\begin{toolbox}{\texttt{embed\_recall}}
\textbf{Description.}\; Semantic search to recall relevant documents from the corpus. Returns a scope file path containing document paths. Use the scope file with \texttt{cat scope\_N.txt | xargs -d '$\backslash$n' rg "pattern"} to search only within recalled documents. May be called multiple times with different queries to create multiple scopes.

\smallskip
\textbf{Parameters.}
\begin{lstlisting}[basicstyle=\ttfamily\scriptsize, escapeinside={}{}, backgroundcolor=\color{white}, frame=none, breaklines=true, columns=fullflexible]
{
  "query": {
    "type": "string",
    "description": "Natural language search query for semantic document recall"
  }
}
// required: ["query"]
\end{lstlisting}
\end{toolbox}

\smallskip
\begin{toolbox}{\texttt{read}}
\textbf{Description.}\; Read the contents of a file. Output is truncated to 2000 lines or 50\,KB (whichever is hit first). Use \texttt{offset}/\texttt{limit} for large files; to read a full file, continue with increasing \texttt{offset} until complete.

\smallskip
\textbf{Parameters.}
\begin{lstlisting}[basicstyle=\ttfamily\scriptsize, escapeinside={}{}, backgroundcolor=\color{white}, frame=none, breaklines=true, columns=fullflexible]
{
  "path":   {"type": "string",  "description": "Path to the file to read (relative or absolute)."},
  "offset": {"type": "integer", "description": "Line number to start reading from (1-indexed)."},
  "limit":  {"type": "integer", "description": "Maximum number of lines to read."}
}
// required: ["path"]
\end{lstlisting}
\end{toolbox}

\smallskip
\begin{toolbox}{\texttt{bash} (standard)}
\textbf{Description.}\; Execute a bash command in the current working directory. Returns stdout and stderr, truncated to the last 2000 lines or 50\,KB (whichever is hit first); the full output is saved to a temp file if truncated. An optional timeout may be provided.

\smallskip
\textbf{Parameters.}
\begin{lstlisting}[basicstyle=\ttfamily\scriptsize, escapeinside={}{}, backgroundcolor=\color{white}, frame=none, breaklines=true, columns=fullflexible]
{
  "command": {"type": "string",  "description": "Bash command to execute."},
  "timeout": {"type": "integer", "description": "Timeout in seconds (optional). rg commands default to 30s."}
}
// required: ["command"]
\end{lstlisting}
\end{toolbox}

\smallskip
\begin{toolbox}{\texttt{bash} (RerankAwareBash variant, \texttt{USE\_CONTEXTUAL\_BASH\_TOOL=1})}
\textbf{Description.}\; Same as the standard \textit{bash} tool, but for \rg~commands an optional \texttt{rerank\_query} may be supplied, used only for semantic reranking of matches.

\smallskip
\textbf{Parameters.}
\begin{lstlisting}[basicstyle=\ttfamily\scriptsize, escapeinside={}{}, backgroundcolor=\color{white}, frame=none, breaklines=true, columns=fullflexible]
{
  "command":      {"type": "string",  "description": "Bash command to execute."},
  "timeout":      {"type": "integer", "description": "Timeout in seconds (optional). rg commands default to 30s."},
  "rerank_query": {"type": "string",  "description":
    "Optional natural-language query used only to rerank rg matches. For rg commands, write exactly two short sentences. Sentence 1 states the final fact, entity, or relation the user is trying to determine from the question. Sentence 2 states what this specific rg step is trying to verify, identify, or retrieve. Build the query by jointly considering the question, the relevant conversation context so far, and the current search step; use concrete entities and make it more specific as the conversation narrows. For non-rg commands, omit this field."}
}
// required: ["command"]
\end{lstlisting}
\end{toolbox}

\subsection{Initial User Prompts}

Each episode begins with a task-specific user prompt that wraps the query. On BC+, it instructs the agent to answer from the local corpus using \rg~and forbids web search; on BRIGHT, it asks the agent to retrieve and rank the relevant documents.

\begin{toolbox}{Initial user prompt on BrowseComp-Plus}
\begin{lstlisting}[basicstyle=\ttfamily\scriptsize, escapeinside={}{}, backgroundcolor=\color{white}, frame=none, breaklines=true, columns=fullflexible]
Answer the following question. The answer is contained in the corpus directory (your current working directory). **Do Not use web search!** Use ripgrep (rg) instead of grep for fast searching.

QUESTION:
{query}
\end{lstlisting}
\end{toolbox}

\smallskip
\begin{toolbox}{Initial user prompt on BRIGHT}
\begin{lstlisting}[basicstyle=\ttfamily\scriptsize, escapeinside={}{}, backgroundcolor=\color{white}, frame=none, breaklines=true, columns=fullflexible]
Retrieve and rank the most relevant documents for the following query.

QUERY:
{query}
\end{lstlisting}
\end{toolbox}


\section{Case Study: Compressed Agent Trajectories}
\label{app:case-study}

\definecolor{rargblue}{HTML}{526D82}
\definecolor{rarggreen}{HTML}{4F786E}
\definecolor{rargorange}{HTML}{A2674D}
\definecolor{querybg}{HTML}{F5F7F8}
\definecolor{queryrule}{HTML}{AEB8BF}

\newcommand{\caseturn}[1]{%
  \begingroup
  \setlength{\fboxsep}{1.3pt}%
  \colorbox{gray!12}{\texttt{\scriptsize #1}}%
  \endgroup}

\newcommand{\queryclue}[1]{%
  \begingroup
  \setlength{\fboxsep}{1.2pt}%
  \colorbox{white}{\strut\scriptsize #1}%
  \endgroup}

We examine a query from the 100K-document BrowseComp-Plus setting.
All three RARG variants are judged correct for the gold answer
\textbf{Russell David Lyons}. This is not a single-hop lookup: the query
requires the agent to join biographical constraints with a three-author paper,
an award, and a second paper title. Tool use decreases strictly across the
variants (33/18/10), making the example a compact view of how relevance
guidance can reduce the interaction needed to resolve a multi-document clue
chain.

\medskip
\noindent
\begin{minipage}{\linewidth}
  \setlength{\fboxsep}{7pt}
  \colorbox{querybg}{%
    \parbox{\dimexpr\linewidth-2\fboxsep\relax}{%
      \textbf{\small Query (Index: 229)}\hfill
      {\scriptsize Gold answer: \textbf{Russell David Lyons}}

      \vspace{5pt}
      \small
      \textit{Find Person A, who earned a Ph.D.\ in Mathematics in 1983 and
      became an AMS Fellow between 2005 and 2020. Person A coauthored a
      1990--2005 paper with Persons B and C; Person B won the Rollo Davidson
      Prize in that period, and Person C published a 1990s paper whose title
      ends in ``Line.'' What is Person A's full name?}
    }%
  }
\end{minipage}
\medskip
We show the three RARG trajectories on a common turn axis in
Figure~\ref{fig:case-trajectory}. For readability, consecutive calls pursuing
the same search objective are manually grouped into a single phase, whereas each
dot denotes the first observation that exposes the full answer-bearing name.
The reported tool totals are computed from the uncompressed trajectories and
therefore include every original call.

\begin{center}
\begin{tikzpicture}[
  x=0.60cm, y=1cm,
  phase/.style={
    rounded corners=1pt,
    minimum height=0.42cm,
    font=\scriptsize,
    inner xsep=3pt
  },
  scope/.style={
    fill=white,
    rounded corners=1pt,
    minimum height=0.42cm,
    font=\scriptsize
  },
  answer/.style={
    fill=white,
    rounded corners=1pt,
    minimum width=1.15cm,
    minimum height=0.42cm,
    font=\scriptsize\bfseries
  }
]
\draw[gray!40] (0,0.90) -- (16,0.90);
\foreach \x in {0,...,16} {
  \draw[gray!40] (\x,0.85) -- (\x,0.95);
  \node[font=\scriptsize, text=gray!70] at (\x,1.12)
    {\number\numexpr\x+1\relax};
}
\node[font=\scriptsize, text=gray!70, anchor=east]
  at (-0.65,1.12) {Turn};

\node[anchor=east, align=left, text width=2.25cm,
  text=rargblue, font=\footnotesize]
  at (-0.65,0)
  {\textbf{\method}\\[-1pt]\scriptsize 17 turns / 33 tools};
\draw[rargblue!45] (0,0) -- (16,0);
\node[scope, draw=rargblue!70] at (-0.2,0) {scope};
\node[
  phase,
  draw=rargblue!70,
  fill=rargblue!11,
  minimum width=3.05cm
] at (3.1,0) {broad clue intersection};
\fill[rargblue] (6,0) circle (2.6pt);
\node[anchor=south, font=\scriptsize] at (6,0.15)
  {\caseturn{T7}~full name exposed};
\node[
  phase,
  draw=rargblue!70,
  fill=rargblue!11,
  minimum width=4.55cm
] at (10.4,0) {coauthor-chain verification};
\node[answer, draw=rargblue] at (16,0)
  {Lyons $\checkmark$};

\node[anchor=east, align=left, text width=2.25cm,
  text=rarggreen, font=\footnotesize]
  at (-0.65,-1.45)
  {\textbf{\methodp}\\[-1pt]\scriptsize 11 turns / 18 tools};
\draw[rarggreen!45] (0,-1.45) -- (10,-1.45);
\node[scope, draw=rarggreen!70] at (-0.05,-1.45) {scope+init};
\fill[rarggreen] (1,-1.47) circle (2.6pt);
\node[anchor=south, font=\scriptsize] at (1.3,-1.30)
  {\caseturn{T2}~full name exposed};
\node[
  phase,
  draw=rarggreen!70,
  fill=rarggreen!11,
  minimum width=3.00cm
] at (4.2,-1.47) {initialized scoped \rg};
\node[font=\scriptsize, text=rarggreen, anchor=north]
  at (8.0,-1.58) {targeted checks};
\node[answer, draw=rarggreen] at (10,-1.45)
  {Lyons $\checkmark$};

\node[anchor=east, align=left, text width=2.25cm,text=rargorange, font=\footnotesize]
  at (-0.65,-2.90)
  {\textbf{\methodpp}\\[-1pt]\scriptsize 10 turns / 10 tools};
\draw[rargorange!45] (0,-2.90) -- (9,-2.90);
\node[scope, draw=rargorange!70] at (-0.05,-2.90) {scope+init};
\fill[rargorange] (1,-2.90) circle (2.6pt);
\node[anchor=south, font=\scriptsize] at (1.35,-2.75)
  {\caseturn{T2}~full name exposed};
\node[
  phase,
  draw=rargorange!70,
  fill=rargorange!11,
  minimum width=3.10cm
] at (4.1,-2.90) {match-reranked scoped \rg};
\node[font=\scriptsize, text=rargorange, anchor=north]
  at (7.3,-3.03) {verify chain};
\node[answer, draw=rargorange] at (9,-2.90)
  {Lyons $\checkmark$};
\end{tikzpicture}
\end{center}

\captionof{figure}{
Compressed, turn-aligned trajectories for BC+ query 229 on the 100K corpus.
The dot indicates the first observation containing the full answer-bearing name.
Initialization moves this evidence from \caseturn{T7} to \caseturn{T2};
all three trajectories are judged correct, while tool use decreases from
33 to 18 to 10.
}
\label{fig:case-trajectory}

\medskip
\noindent\textbf{How the search state changes.}
The trajectory lengths alone do not explain the difference among the methods.
The key distinction is the information state created by the first decisive
observation. Table~\ref{tab:case-decisive-evidence} shows the earliest evidence
that turns the task from an open-ended person search into verification of a
specific mathematician and coauthor chain.

\begin{table}[h]
  \centering
  \footnotesize
  \setlength{\tabcolsep}{6pt}
  \renewcommand{\arraystretch}{1.35}
  \caption{
  Earliest observation that exposes the full answer-bearing name and its
  immediate effect on the subsequent search strategy.
  }
  \label{tab:case-decisive-evidence}
  \begin{tabular}{
    @{}
    p{0.08\linewidth}
    p{0.4\linewidth}
    p{0.43\linewidth}
    @{}
  }
    \toprule
    \textbf{Method}
    &
    \textbf{First decisive observation}
    &
    \textbf{Resulting search transition}
    \\
    \midrule

    \textcolor{rargblue}{\textbf{\method}}
    &
    \caseturn{T7}\quad
    \textit{``Russell David Lyons \ldots{} Ph.D., August 1983,
    Mathematics''}
    &
    After broad searches over prize winners and possible coauthors, the CV
    identifies Person A; later turns establish the Peres--Pemantle paper chain.
    \\

    \addlinespace[3pt]

    \textcolor{rarggreen}{\textbf{\methodp}}
    &
    \caseturn{T2}\quad
    \textit{``Russell David Lyons \ldots{} Ph.D., August 1983,
    Mathematics''}
    &
    Initialization places the CV in the first scoped searches; the remaining
    interaction verifies the award, coauthorship, and title-ending clue.
    \\

    \addlinespace[3pt]

    \textcolor{rargorange}{\textbf{\methodpp}}
    &
    \caseturn{T2}\quad
    \textit{``Russell David Lyons \ldots{} Ph.D., August 1983,
    Mathematics''}
    &
    The match-reranked scope exposes the same CV in a single post-initialization
    call; one-tool turns then verify each remaining link.
    \\

    \bottomrule
  \end{tabular}
\end{table}

\noindent\textbf{Case takeaway.}
All three trajectories are judged correct for Russell David Lyons despite the
query's multi-document joins. Tool use decreases strictly from 33 to 18 to 10,
and total turns fall from 17 to 11 to 10. Initialization accounts for the
largest change in evidence timing, moving the answer-bearing CV from
\caseturn{T7} to \caseturn{T2}; match-level reranking retains that early exposure
while further reducing tool use. Rather than implying a general ordering from a
single case, this example illustrates how progressively finer relevance
guidance can compress the interaction required to verify a difficult clue
chain.

\end{document}